\documentclass[conference]{IEEEtran}
\IEEEoverridecommandlockouts
\usepackage{cite}
\usepackage{amsmath,amssymb,amsfonts}
\usepackage{algorithmic}
\usepackage{graphicx}
\usepackage{textcomp}
\usepackage{xcolor}
\usepackage[caption=false,font=footnotesize]{subfig}
\usepackage{float}
\usepackage{makecell}
\usepackage{multirow}
\usepackage{hyperref}
\usepackage[flushleft]{threeparttable}

\def\BibTeX{{\rm B\kern-.05em{\sc i\kern-.025em b}\kern-.08em
    T\kern-.1667em\lower.7ex\hbox{E}\kern-.125emX}}

\DeclareRobustCommand{\IEEEauthorrefmark}[1]{\smash{\textsuperscript{\footnotesize #1}}}

\begin{document}

\title{Design of a Breakaway Utensil Attachment for Enhanced Safety in Robot-Assisted Feeding
\\
\thanks{
The research was conducted at the Future Health Technologies at the Singapore-ETH Centre, which was established collaboratively between ETH Zurich and the National Research Foundation Singapore. This research is supported by the National Research Foundation Singapore (NRF) under its Campus for Research Excellence and Technological Enterprise (CREATE) program.
}

\thanks{
\IEEEauthorrefmark{1}School of Mechanical and Aerospace Engineering, Nanyang Technological University Singapore
}
\thanks{
\IEEEauthorrefmark{2}Singapore-ETH Centre, Future Health Technologies Programme
}
\thanks{
Corresponding author's email: \texttt{janne.yow@ntu.edu.sg}}

}

\author{
    \IEEEauthorblockN{
        Hau Wen Chang\IEEEauthorrefmark{1}, 
        J-Anne Yow\IEEEauthorrefmark{1,}\IEEEauthorrefmark{2}, 
        Lek Syn Lim\IEEEauthorrefmark{1}
        Wei Tech Ang\IEEEauthorrefmark{1,}\IEEEauthorrefmark{2}
    }
}

\maketitle

\begin{abstract}
Robot-assisted feeding systems enhance the independence of individuals with motor impairments and alleviate caregiver burden. While existing systems predominantly rely on software-based safety features to mitigate risks during unforeseen collisions, this study explores the use of a mechanical fail-safe to improve safety. We designed a breakaway utensil attachment that decouples forces exerted by the robot on the user when excessive forces occur. Finite element analysis (FEA) simulations were performed to predict failure points under various loading conditions, followed by experimental validation using 3D-printed attachments with variations in slot depth and wall loops. To facilitate testing, a drop test rig was developed and validated.
Our results demonstrated a consistent failure point at the slot of the attachment, with a slot depth of 1 mm and three wall loops achieving failure at the target force of 65 N. Additionally, the parameters can be tailored to customize the breakaway force based on user-specific factors, such as comfort and pain tolerance. CAD files and utensil assembly instructions can be found here: \url{https://tinyurl.com/rfa-utensil-attachment}

\end{abstract}



\section{Introduction}
Robot-assisted feeding systems offer a promising solution for improving the quality of life of those requiring feeding assistance and alleviating the burden on caregivers. Despite significant advancements in developing robust, intelligent, and personalized robot-assisted feeding systems \cite{park2020active, gordon2024adaptable}, a notable gap exists in designing mechanical fail-safes to ensure user safety. One of the critical safety concerns is the excessive force that can be applied during feeding due to the close physical human-robot interaction required for the task, which poses risks such as discomfort, injury, or damage to the user's oral cavity. 

To address this, designing robot systems that limit the force applied during interaction is paramount. Current research systems often focus on sophisticated control algorithms and sensor feedback mechanisms to reduce force \cite{shaikewitz2023mouth, jenamani2024feel}; however, mechanical fail-safes provide an additional layer of protection, particularly in unforeseen scenarios where software-controlled mechanisms may not respond. A mechanical solution offers a passive, reliable response to excessive force without relying on the system's active feedback loops.

This work presents a breakaway utensil attachment designed for robot-assisted feeding. The mechanism ensures that when the robotic arm applies excessive force on the user, the spoon automatically detaches, effectively preventing further force transmission to the user. By incorporating this passive fail-safe, we aim to enhance user safety and provide a safeguard against potential harm in unpredictable scenarios.

\section{Related Work}
\subsection{Utensil Attachments for Robot-Assisted Feeding}
Grasping a utensil directly using a two-finger gripper, commonly found on robot manipulators, is challenging. To address this, utensil attachments are typically employed to enable secure grasping by connecting utensils, such as spoons or forks, to structures that are easier for the robot to handle \cite{park2020active, gordon2024adaptable, belkhale2022balancing, shaikewitz2023mouth, jenamani2024feel}. Additionally, force-torque (FT) sensors are often integrated into these attachments to provide haptic feedback, enabling more precise control during feeding tasks. 

Recently, more innovative utensil designs have been proposed that do not take the form of a traditional spoon or fork -- Kiri-spoon \cite{keely2024kiri} uses a soft kirigami structure, allowing it to encapsulate and release food easily during bite acquisition. Separately, Gordon \textit{et. al.} \cite{gordon2024adaptable} presented their robot-assisted feeding system, which features a utensil attachment with a mechanically engineered weak point. This weak point is designed to break when excessive force is applied, eliminating any physical interaction between the robot and the user. However, the design specifics of this mechanism are not detailed, and there remains limited research exploring the integration of mechanical fail-safes in robot-assisted feeding systems.

\subsection{Safety in Assistive Robotics}
Safety in human-robot interaction is paramount, especially in assistive robotics, where physical human-robot interaction is common, and robots operate in close proximity to vulnerable users. Existing research on safety can be categorized into three main categories -- control methods, motion planning, and behavior prediction. Approaches relying on control methods typically use force-torque sensors, impedance control, or compliance-based control to detect and mitigate excessive forces during interaction \cite{shaikewitz2023mouth, jenamani2024feel}. Motion planning approaches aim to preemptively avoid collisions by generating safe trajectories based on real-time sensing and constraints \cite{lasota2014toward}. However, contact is unavoidable in physically assistive tasks such as feeding. For behavior prediction, the focus is on predicting human intent and motion \cite{lasota2015analyzing, zhao2019walking}, enabling the robot to adapt its actions proactively to prevent potentially unsafe interactions.

While these methods effectively reduce safety risks, they primarily rely on software-based solutions and AI-driven models, which can be limited by sensor inaccuracies, computational latency, and unanticipated failure modes. Furthermore, these approaches often depend on complex system integrations \cite{hamad2023concise} that may not respond adequately in every scenario. In contrast, our work introduces a safety feature based on mechanical principles, providing an additional safety layer in assistive robotic systems.

\begin{figure}[t]
    \centering
    \subfloat[]{%
        \includegraphics[height=5cm]{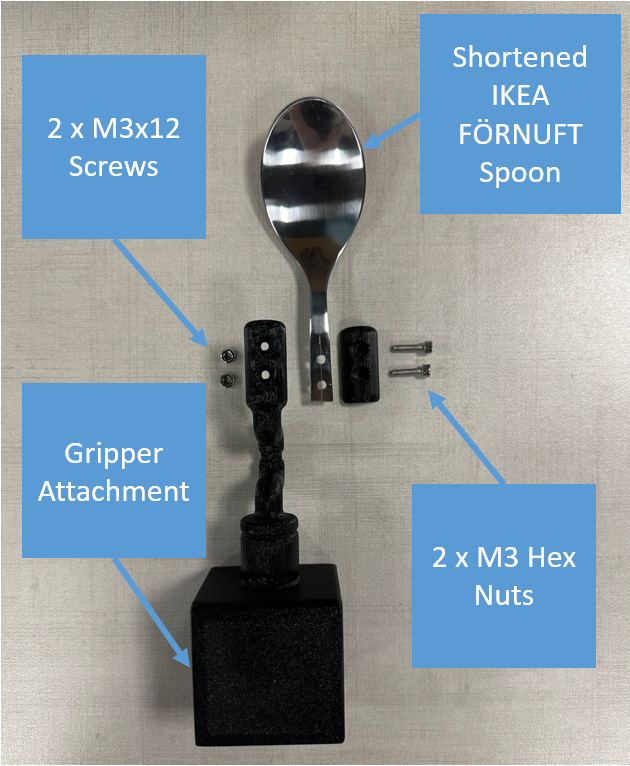}
        \label{fig: utensil-attachment}
    }
    \hfill
    \subfloat[]{%
        \includegraphics[height=5cm]{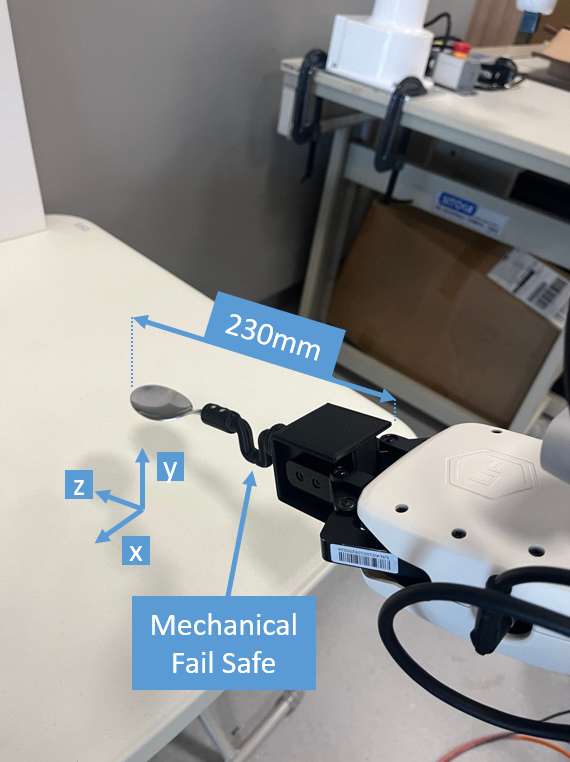}
        \label{fig: arm-with-utensil-attachment}
    }
    \caption{Utensil attachment with a mechanical failsafe (a) Components of our proposed utensil attachment (b) Utensil attachment grasped by a two-finger gripper}
\end{figure}

\section{Approach}
\subsection{Design Considerations and Constraints}

We designed a utensil attachment for the IKEA F\"{o}rnuft spoon, ensuring compatibility with any robot manipulator with a two-finger gripper. To secure the spoon, its handle was shortened and attached to the utensil holder using two M3 screws and nuts, as shown in Fig. \ref{fig: utensil-attachment}. The utensil and attachment's total length was limited to a maximum of 230mm to maintain the robot's range of motion during feeding and prevent interference with its movements of the surrounding environment (Fig. \ref{fig: arm-with-utensil-attachment}).

In assistive feeding, excessive force can occur in two scenarios: (1) sudden impact collisions and (2) slow compressive forces. Sudden impact collisions, such as an unintended rapid movement of the arm striking the user, pose a greater hazard as they can deliver high, localized forces in a very short time frame, increasing the risk of injury. Thus, this work focuses on addressing sudden impact collisions. The mechanical fail-safe was optimized to break at an impact force of $F_{\text{impact}} = 65N$, based on the maximum permissible force that collaborative robots can safely exert on a human face, as defined by ISO 15066 \cite{ISO15066}, which details safety requirements for collaborative robots.

Additionally, the utensil attachment must be robust enough to withstand forces encountered during normal feeding operations, such as acquiring food and transferring it to the user's mouth. To determine the minimum force threshold, we recorded force-torque data while scooping uncooked black beans, representing a high-force food scenario. To account for variability in food textures, we applied a safety factor of 2.5, resulting in a minimum force threshold of $F = 25N$. This balance between breakaway safety and functional durability is critical to the attachment's performance in real-world use.


\subsection{Breakaway Mechanism Design}
Standard breakaway mechanisms like shear pins, slip clutches, and magnetic couplings were initially considered but introduced bulk, complexity, and higher costs. Instead, we integrated the failure point directly into the 3D-printed shaft, eliminating additional components while maintaining a compact design. We achieved controlled failure by leveraging force moments and reducing the cross-sectional diameter, amplifying stress within the part. Localized stress concentration features, such as notches and slots, were introduced to create predictable failure points, ensuring a consistent breakaway force while maintaining a compact and lightweight design.

The reduced diameter \(d\) in these regions increases both torsional stress \(\tau\) and bending stress \(\sigma\), as described by:
\begin{equation}
    \tau = \frac{16T}{\pi d^3},
    \sigma = \frac{32M}{\pi d^3}
\end{equation}
where \(T\) is the applied torque, \(M\) is the bending moment, and 
\(d\) is the diameter of the shaft. These stresses combine to produce higher equivalent (Von-Mises) stresses \(\sigma_{e}\), which predict failure under complex loading conditions. Von-Mises stress is calculated as:
\begin{equation}
    \sigma_{e} = \sqrt{\sigma_{x}^2+\sigma_{y}^2+\sigma_{x}^2\sigma_{y}^2-3\tau_{xy}^2}
\end{equation}
where \(\sigma_{x}\) and \(\sigma_{y}\) are the normal stresses in the x and y directions, and \(\tau_{xy}\) is the shear stress in the x-y plane. Von-Mises stress is used in Section \ref{section FEA} to analyze how the stress is distributed under different loading conditions (Fig. \ref{fig: FEA}).

\begin{figure}[t]
    \centering
    \subfloat[]{%
        \includegraphics[height=3.2cm]{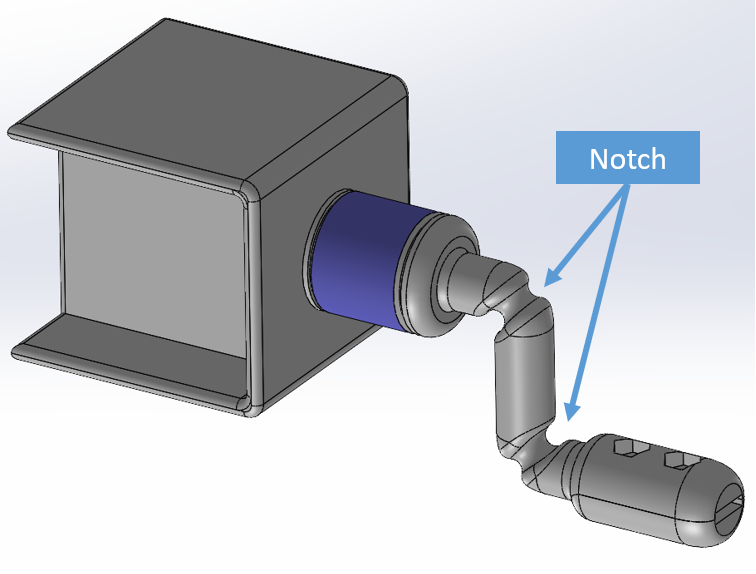}
        \label{fig: L-shape design}
    }
    \hfill
    \subfloat[]{%
        \includegraphics[height=3.2cm]{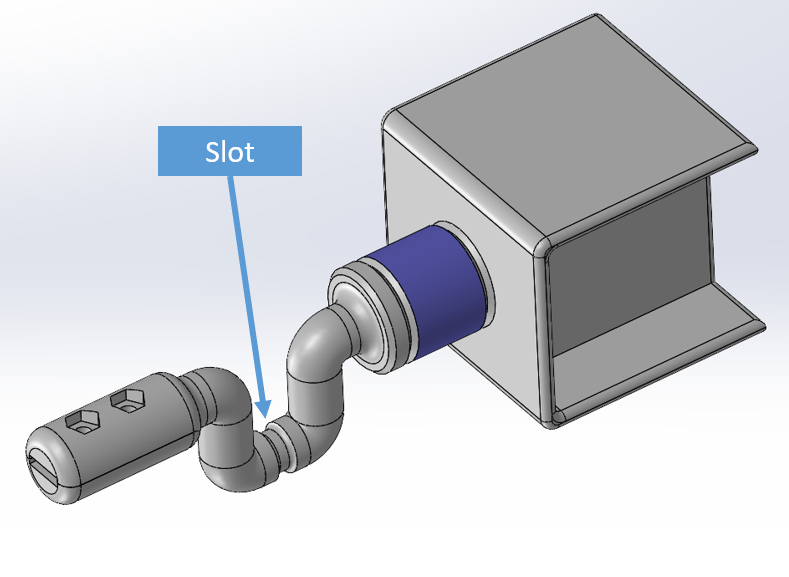}
        \label{fig: U-shape design}
    }
    \caption{Utensil attachment designs -- (a) L-shaped design with notches at both corners (b) U-shaped design with a slot in the horizontal section}
    \label{fig: attachment-designs}
\end{figure}

Guided by these principles, two breakaway mechanism designs were developed: (1) an L-shaped design with notches at the corners and (2) a U-shaped design with a slot in the horizontal section (Fig. \ref{fig: attachment-designs}). These features ensure failure occurs predictably at the intended points under applied forces.

\subsection{Finite Element Analysis (FEA)}
\label{section FEA}

Finite Element Analysis (FEA) was conducted using SolidWorks to evaluate the stress distribution and predict the failure points for each design under a compressive force of $F_{\text{compressive}}=65N$, corresponding to the ISO 15066 \cite{ISO15066} threshold. 
Regions with high-stress concentrations were identified as likely failure points under accidental impacts.

To simulate real-world conditions, FEA was performed with different ends of the part fixed, accounting for variations in how the utensil attachment might interact with the robot’s end-effector or contact the user. This provided a comprehensive evaluation of stress behavior. We focused on front-facing impacts, as they are the most common during feeding, with the robot and spoon approaching the user from the front. 

As the parts were designed for 3D printing, anisotropic strength properties resulting from the layer-by-layer construction process introduced challenges in accurately simulating their performance \cite{gibson2021additive, forster2015materials}. Factors such as print orientation, infill pattern, and wall thickness influence strength, making it difficult for FEA to provide exact predictions \cite{abbot2019finite}. Therefore, the FEA results were treated as qualitative guides for identifying potential failure points rather than definitive performance metrics.

\begin{figure}
    \centering
    \subfloat[]{%
        \includegraphics[height=4.5cm]{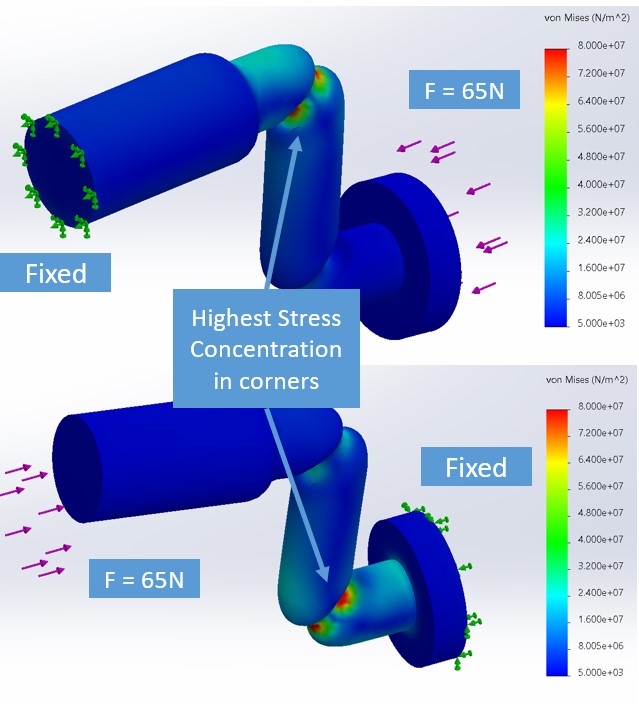}
        \label{fig: L-shape FEA}
    }
    \hfill
    \subfloat[]{%
        \includegraphics[height=4.5cm]{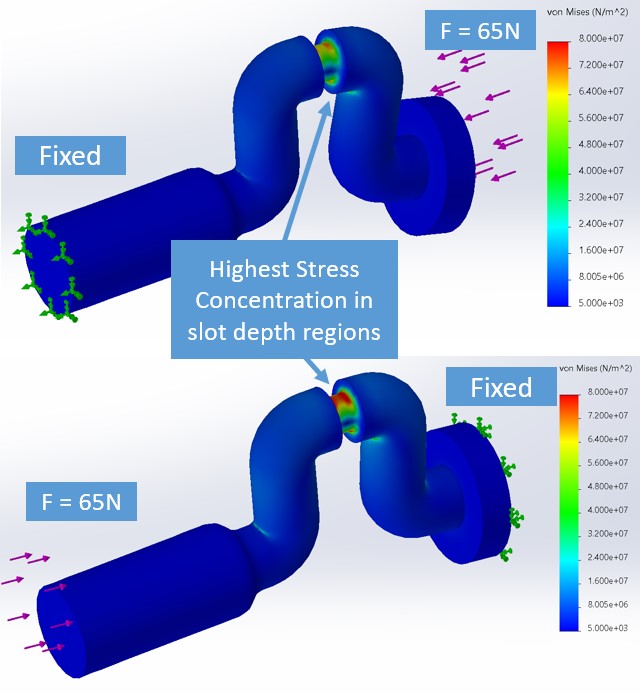}
        \label{fig: U-shape FEA}
    }
    \caption{Finite element analysis (FEA) on the L-shaped (left)o and U-shaped (right) design. (Top) FEA where the spoon end is fixed and (bottom) FEA where the opposite end is fixed.}
    \label{fig: FEA}
\end{figure}

FEA results on the L-shaped design revealed that stress concentrations were highest at the corners of the bend (Fig. \ref{fig: L-shape FEA}). However, the specific failure location varied based on which end of the part was fixed. This made predicting the exact failure location challenging, complicating systematic testing and parameter adjustments.

In contrast, the U-shaped design demonstrated consistent stress concentrations at the slot, resulting in a single predictable failure point (Fig. \ref{fig: U-shape FEA}). Given its predictability, the U-shaped design was ultimately selected. The slot depth \(d\) was identified as a key variable for systematically evaluating breakage forces.

\subsection{Fabrication via 3D printing}
\begin{figure}[t]
    \centering
    \subfloat[]{%
        \includegraphics[height=5.5cm]{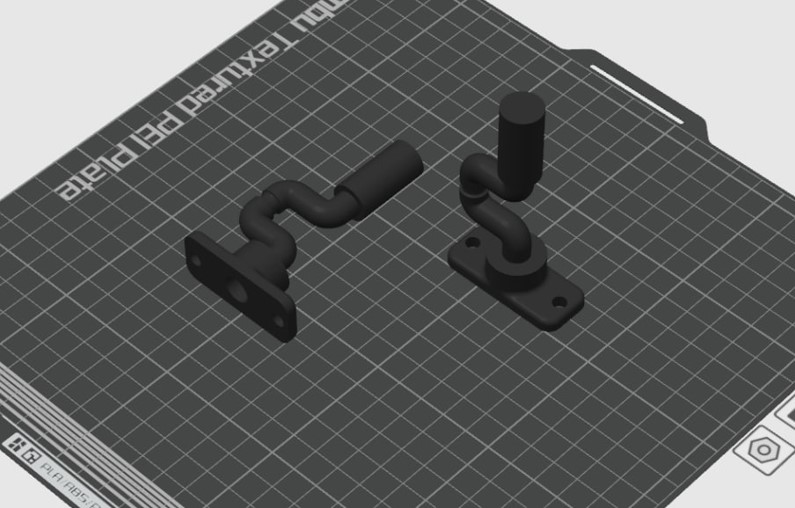}
        \label{fig: Bambu print orientation}
    }
    \hfill
    \subfloat[]{%
        \includegraphics[height=4.4cm]{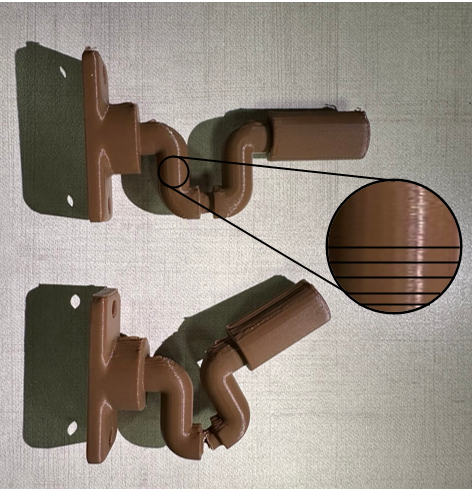}
        \label{fig: Conventional print orientation}
    }
    \hfill
    \subfloat[]{%
        \includegraphics[height=4.4cm]{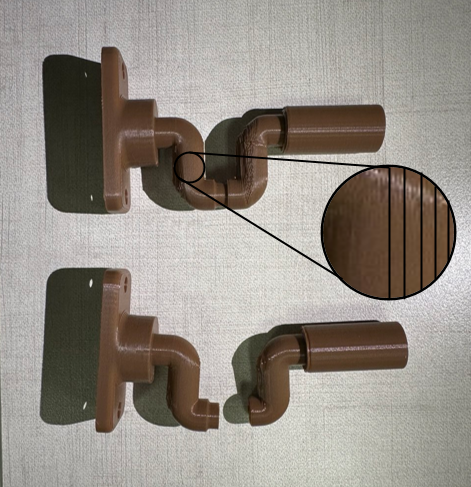}
        \label{fig: Unconventional print orientation}
    }
    \caption{3D printing orientations -- (a) Visualization of two 3D printing orientations in the Bambu slicer: the left corresponds to (b), and the right corresponds to (c). (b) Print orientation with layer lines perpendicular to the breaking line for strength, resulting in an unclean break (c) Print orientation with layer lines parallel to the breaking line, producing a clean break}
    \label{fig: Print orientation}
\end{figure}

To facilitate rapid prototyping and iterative testing, the utensil attachments were 3D printed. Beyond varying the slot depth \(d\), 3D printing parameters such as wall thickness \(w\) and print orientation (Fig. \ref{fig: Bambu print orientation}) were adjusted to control the strength and performance of each part. A 15\%-density grid pattern was used as the infill for all printed parts. All parts were printed using the Bambu Lab X1C 3D printer (China) and the eSun PLA+ 1.75mm filament (China) to ensure uniformity across iterations and eliminate variability from printer inconsistencies.

Print orientation was critical due to the anisotropic strength of 3D-printed parts. Initially, parts were oriented with layer lines perpendicular to the intended breaking line to maximize strength \cite{kelecs2017effect, kovan2017effect}. However, this configuration resulted in inconsistent and unclean breaks (Fig. \ref{fig: Conventional print orientation}), making it more challenging to predict the failure behavior.

To address this, the layer lines were aligned parallel to the breaking line (Fig. \ref{fig: Unconventional print orientation}). This orientation leveraged the inherent layer-to-layer weakness of 3D printing, ensuring predictable failure at the desired location. The mechanical failure point resulting from a collision produces a clean break with smooth edges, minimizing the risk of injury to the patient. Wall thickness \(w\) was also varied to fine-tune the breakaway force, with thicker walls increasing strength and thinner walls reducing the force.

\section{Experiments}

\subsection{Experimental Setup and Design}
\begin{figure}
    \centering
        \includegraphics[width=0.6\linewidth]{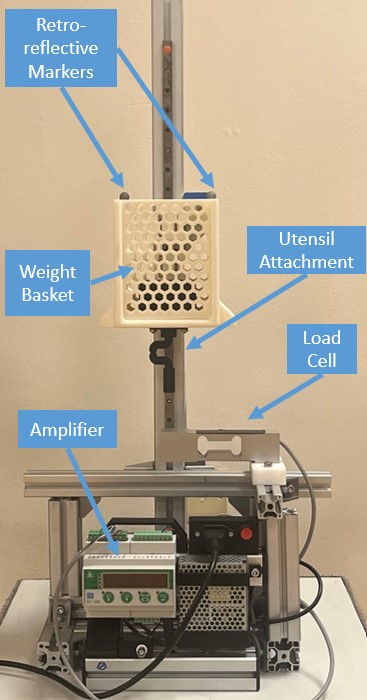}
    \caption{Drop test rig}
    \label{fig: collision rig updated}
\end{figure}

A drop test rig was developed to evaluate the impact strength of the utensil attachments (Fig. \ref{fig: collision rig updated}). The rig included a 20kg single-point load cell with an intrinsic resolution of 1/3000 of its full-scale range (SIWAREX WL260 SP-S SA, Siemens, Germany) to measure the impact forces. Data acquisition (DAQ) was facilitated using a 24-bit amplifier (DAT 1400, Pavone Systems, Italy), which amplified and conditioned the signals from the load cell. The system was calibrated using the Optimation software (Pavone Systems, Italy) with a scaling factor of 1V = 20N for voltage-to-force conversion. The amplifier's display resolution was set to \(\pm 0.01N\).

A motion capture system (Qualisys AB, Sweden) was employed to record the load cell readings and the kinematics of each drop, including position, velocity, and acceleration. Retro-reflective optical markers were placed on the basket of the test rig for precise motion tracking. The Qualisys Track Manager (QTM) v2019 was used as an integrated software interface for seamless and easy-to-use data recording, which we used to synchronize the cameras with the load cell. The trajectories of the markers and the load cell readings were captured synchronously at 200Hz and 2000Hz, respectively.




To ensure consistent alignment during testing, a 500mm linear guide rail with two linear guide blocks stabilized the weight basket. Grease was applied to the rail interface to reduce friction and ensure smooth motion. The weight basket was designed to hold varying masses, and the test parts were attached to it for drop testing. 

The drop height was controlled using a 3D-printed slot and a brass rod mechanism. The slot, mounted on the central aluminum profile, held the brass rod in place until manually released. 
This setup ensured consistent and reliable impact force readings at each height.


The rig's framework was constructed using 40 x 40mm aluminum profiles and secured with L-brackets to minimize structural damping during collisions. Additionally, two 10kg weights were placed on the rig to enhance stability and further reduce damping effects.




\subsection{Validation of Test Rig}
Before conducting our experiments, we conducted some validation checks. Static validation was performed by loading calibrated weights onto the load cell and verifying that the readings matched the known weights. The measured readings were consistent with the known weights, confirming the static accuracy of the load cell.

The actual impact force \(F_{actual}\) was determined using the peak voltage \(V_{peak}\) from the voltage-time graph recorded by the load cell. With a scaling factor of 1V/20N, the impact force was calculated as:

\begin{equation}
F_{\text{impact}}=(\frac{20N}{V})V_{\text{peak}}\label{eq}
\end{equation}

The theoretical impact force \(F_{theoretical}\) was calculated based on the principle of energy conservation, assuming no energy losses during impact. The kinetic energy of the weight basket and utensil attachment before impact \(KE\) was equated to the work done \(WD\) on the load cell:

\begin{equation*}
    KE = \frac{1}{2}mv^2, WD=Fd_{\text{stop}}, KE=WD
\end{equation*}

From these equations, the impact force was derived as:
\begin{equation}
    F_{theoretical}=\frac{mv^2}{2d_{\text{stop}}}
\end{equation}
where $m$ is the total mass of the weight basket and attachment, $v$ is max velocity before impact, and $d_{\text{stop}}$ is the stopping distance, approximated as the difference between the resting position \(p_{rest}\) and the lowest position \(p_{lowest}\), i.e. \(d_{\text{stop}} = p_{\text{rest}} - p_{\text{lowest}}\). The lowest position is determined by identifying the lowest point on the displacement vs. time graph, while the resting position is obtained from the initial state when the weight basket rests on the load cell.

\begin{table}[htbp]
\caption{Sample Values for Impact Force Validation}
\begin{center}
\begin{tabular}{|c|c|}
\hline
\textbf{Variable}                     & \textbf{Value} \\ \hline
Mass, $m$                                   & 0.735 kg       \\ \hline
Resting Position, $p_{\text{rest}}$      & 690.489 mm     \\ \hline
Lowest Position, $p_{\text{lowest}}$        & 687.429 mm     \\ \hline
Max Velocity, $v$                           & 860.634 mm/s   \\ \hline
Theoretical Impact Force, $F_{\text{theoretical}}$ & 89.0 N  \\ \hline
\hline
Peak Voltage, $V_{\text{peak}}$             & 3.78 V         \\ \hline
Actual Impact Force, $F_{\text{actual}}$    & 75.6 N         \\ \hline
\hline
Percentage Error, $e$                       & 17.7\%           \\ \hline
\end{tabular}
\label{tab: sample-calc}
\end{center}
\end{table}

The sample values used in the validation are summarized in Table \ref{tab: sample-calc}.
As expected, the theoretical impact force was greater than the experimental value due to the assumption of no energy losses in the theoretical calculation. In real-life experiments, energy losses occur due to factors such as sound generation and damping in the rig structure. 
The difference between the theoretical and actual impact forces of \(13.4 N\) corresponds to an error of \(17.7\%\), which is within the acceptable range, validating the rig for further testing.

\subsection{Experiment Procedure}
In our experiments, we evaluated the strength of the utensil attachment by varying two parameters: the slot depth \(d\) and the number of wall loops \(w\) used during 3D printing. To determine the minimum force required to break the attachment, we conducted drop tests by varying the height at which the weight basket and attachment were released onto the load cell. If the attachment did not break, we recorded the peak forces and incrementally increased the drop height by \(1.0cm\) until the attachment broke.

Once the approximate range of heights where the part transitioned between breaking and not breaking, we refined our testing by reducing the height increments to \(0.2cm\). Height increments were limited to \(0.2cm\) during this stage, as smaller increments of \(0.1cm\) were impractical to achieve consistently. For instance, if the part broke at \(h=5.0cm\) but did not break at \(h=4.0cm\), we tested intermediate heights such as \(h=4.8cm\) and \(h=4.6cm\). Starting from the highest height within the identified range, we conducted three trials at each height to ensure consistency and reliability of the results. 

\begin{figure}[htbp]
    \centering
    \subfloat[]{%
        \includegraphics[width=0.45\textwidth]{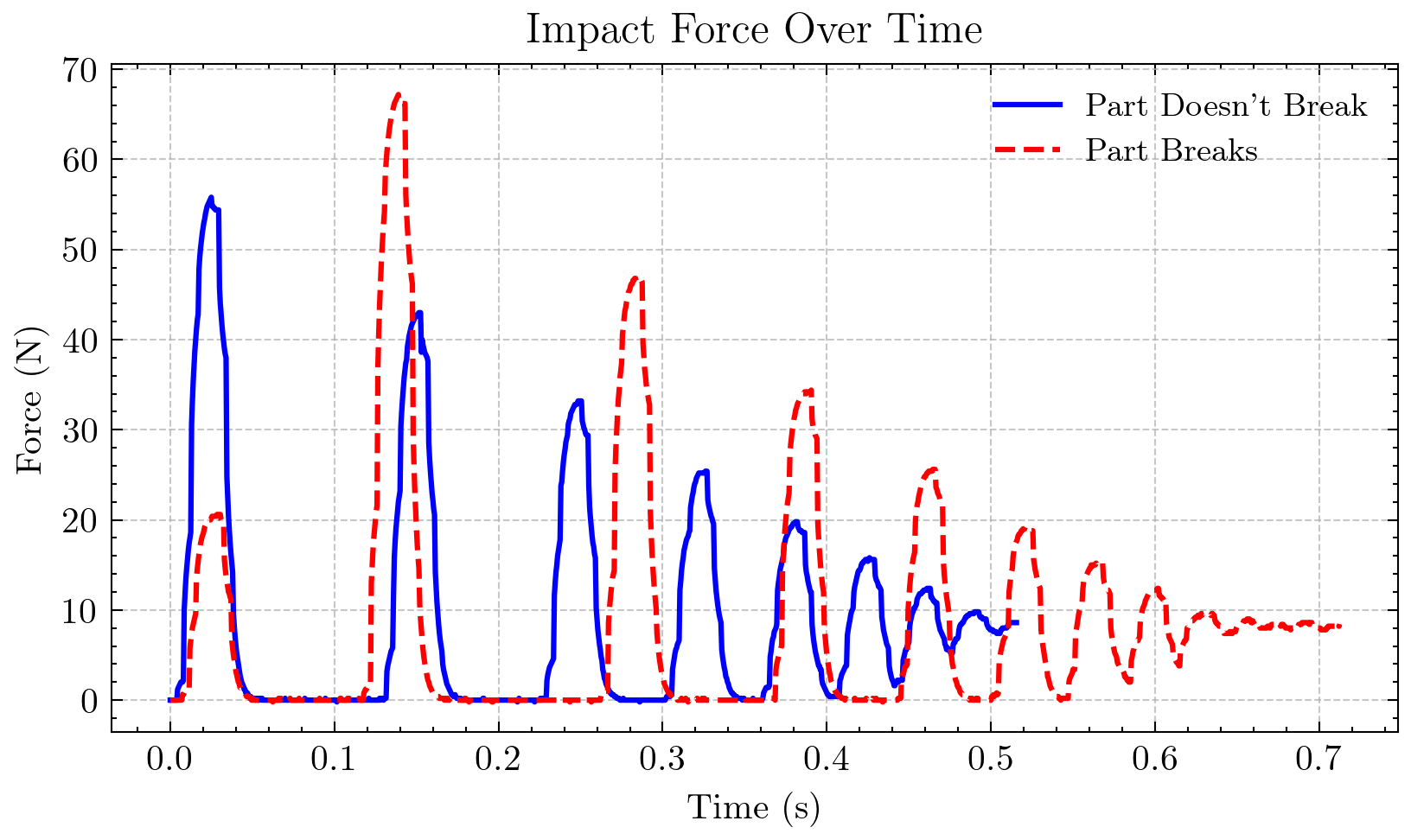}
        \label{fig: force-plot-unbroken}
    }
    \hfill
    \caption{Force-time graph for a part that does not break (blue solid line) and a part that breaks (red dashed line). The impact force is determined from the primary peak in the blue solid line, representing the maximum impact force experienced by the part. In contrast, for the red dashed line, the impact force cannot be derived due to the smaller first peak, which occurs as the part breaks upon impact.}
\end{figure}

If the part consistently broke at a specific height, we decreased the height incrementally and repeated the trials until the parts stopped breaking. The impact force was the primary peak force in the force-time graph for parts that did not break (Fig. \ref{fig: force-plot-unbroken}). However, for parts that broke, the impact force could not be derived from the graph due to the smaller first peak, as energy is dissipated through the breaking process rather than being fully transferred to the load cell (Fig. \ref{fig: force-plot-broken}). Thus, the breaking height was the highest height where the part did not consistently break across trials. For example, if the part consistently broke at \(h=4.6cm\) but started not breaking at \(h=4.4cm\), we considered \(h=4.4cm\) as the breaking height as shown in Table \ref{tab:impact-force}. 

Once the breaking height was determined, we calculated the average peak force recorded during the trials at that height, where some or all parts did not break, to estimate the breaking force. This approach provided a conservative estimate of the breaking force, reflecting the force just below the threshold of consistent failure. Hence, the actual impact force required to break the attachment is slightly higher than the average impact force recorded.

\begin{table}[htbp]
\caption{Peak Impact Forces for \(d=1.0mm\) and \(w=3\)}
\begin{center}
\begin{threeparttable}
    \begin{tabular}{|c|c|c|c|c|}
    \hline
    Height (cm) & $t_1$ (N) & $t_2$ (N) & $t_3$ (N) & Average (N) \\ \hline
    4.2 & 62.8 & 63.4 & 63.4 & 63.2 \\ \hline
    4.4 & N/A  & 65.0 & N/A  & 65.0 \\ \hline
    4.6 & N/A  & N/A  & N/A  & N/A  \\ \hline
    4.8 & N/A  & N/A  & N/A  & N/A  \\ \hline
    5.0 & N/A  & N/A  & N/A  & N/A  \\ \hline
    \end{tabular}
    \begin{tablenotes}
        \item{N/A indicates that part broke during testing}
    \end{tablenotes}
\end{threeparttable}
\end{center}
\label{tab:impact-force}
\end{table}

\begin{table}[t]
    \caption{Effect of Slot Depth $d$ and Wall Loop $w$ on Impact Force}
    \begin{center}
    \begin{tabular}{|c|c|c|}
    \hline
    \thead{Slot Depth\\$d$ (mm)} & \thead{Wall Loops\\$w$} & \thead{Average\\Peak Force\\ $F_{\text{avg}}$ (N)} \\ \hline
    1.0 & 6 & 75.6 \\ \hline
    1.0 & 3 & 65.0 \\ \hline
    2.0 & 6 & 53.1 \\ \hline
    2.0 & 3 & 45.0 \\ \hline
    \end{tabular}
    \label{tab:slot_depth_forces}
    \end{center}
\end{table}

\begin{figure}[htbp]
    \centerline{\includegraphics[width=0.32\textwidth]{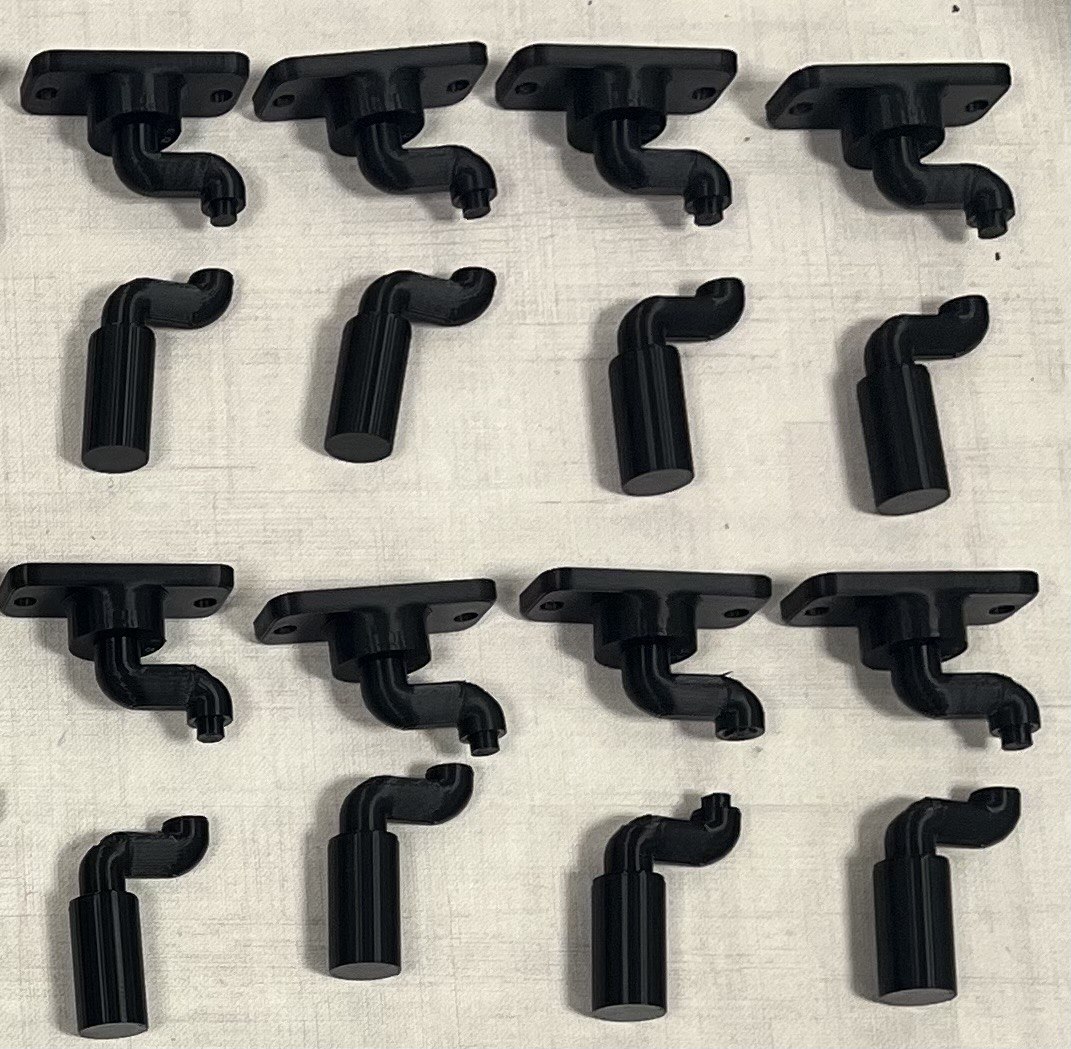}}
    \caption{Consistent failure point at the slot of the breakaway mechanism}
    \label{fig: consistent failure}
\end{figure}

\section{Results}

The results from the experiments are summarized in Table \ref{tab:slot_depth_forces}. Based on the data, we observe that the impact force required to break the attachment increases with the slot depth \(d\) and decreases with the number of wall loops \(w\).

The configuration that produces an impact force closest to the maximum permissible force \(F_{\text{max}} = 65N\) is achieved when the attachment has a slot depth of \(d = 1mm\) and is printed with \(w = 3\) wall loops, which yielded an average impact force of \(F_{\text{impact}} = 65.0N\). This configuration is selected as the optimal design for the breakaway mechanism.

In scenarios where the maximum permissible force needs to be reduced -- for example, to accommodate lower pain thresholds or to prioritize enhanced safety -- the parameters can be refined to achieve the desired reduction. For instance, if a maximum permissible force of \(F_{\text{max}} = 50N\) is desired, the configuration with \(d=2.0mm\) and \(w=6\) wall loops, which produces an average impact force of \(F_{impact} = 53.1N\), could be adjusted by reducing the number of wall loops to lower the breaking force further.

These results demonstrate the utility of systematically varying slot depth and wall loop parameters to design a breakaway mechanism that can be tailored to meet specific force requirements. Furthermore, our experiments revealed that our design resulted in a consistent failure point, which is at the slot of the utensil attachment (Fig. \ref{fig: consistent failure}).





\section{Conclusion and Future Work}

In conclusion, we designed a utensil attachment with a mechanical fail-safe that breaks around the maximum permissible force exerted on a human face by robots $F_{\text{max}} = 65N$. To ensure the part breaks at a consistent breaking point, the parts were printed with the layer lines parallel to the intended breaking line, allowing for controlled and predictable failure behavior. To evaluate the performance of the mechanism, we developed a drop test rig, simulating collisions between the robot arm and the user and conducted systematic experimental testing. Through our experiments, we established how design parameters such as slot depth \(d\) and wall loops \(w\) affect the impact force required to break the attachment.

While the results offer valuable insights into designing effective mechanical fail-safes, the study has some limitations. For instance, the testing procedure did not account for impacts at angles, which could affect the breaking behavior. The experiments were also conducted in a controlled setting using a drop test rig, which does not fully replicate real-world interactions between the utensil and user.

Future work should address these limitations by testing angled impacts and testing with the actual robotic arm and utensil to evaluate performance under real-world conditions. Additionally, other design and 3D printing parameters, such as infill density and infill pattern, could also be investigated to optimize the attachment further. Furthermore, determining the maximum permissible force based on user-specific factors, such as pain tolerance and comfort levels \cite{han2022assessment}, could further enhance the safety and customization of these attachments.

\bibliographystyle{IEEEtran}
\bibliography{IEEEabrv, references}

\end{document}